\RequirePackage[hyphens]{url}
\documentclass[letterpaper, 10pt, conference]{ieeeconf}      
\IEEEoverridecommandlockouts                              
\usepackage{amsmath,bm}                             
\usepackage{array}
\usepackage{amssymb,amsfonts}
\usepackage{booktabs}
\usepackage[all,arc]{xy}
\usepackage{enumerate}
\usepackage{mathrsfs}
\usepackage{titletoc}
\usepackage[noend]{algpseudocode}
\usepackage{graphicx}
\usepackage[caption=false,font=footnotesize]{subfig}
\usepackage{balance}
\usepackage{algorithm}
\usepackage[letterpaper, left=0.75in, right=0.75in, bottom=0.75in, top=1.0in]{geometry}
\usepackage{cite}
\usepackage{array}
\usepackage{url}
\usepackage{multirow}
\usepackage[table,xcdraw]{xcolor}
\usepackage[utf8]{inputenc}
\usepackage{mathtools}
\PassOptionsToPackage{hyphens}{url}\usepackage{hyperref}

\title{\LARGE \bf 
A Survey of Deep Reinforcement Learning Algorithms for \\Motion Planning and Control of Autonomous Vehicles}

\author{Fei Ye$^1$, Shen Zhang$^2$, Pin Wang$^{3}$, and Ching-Yao Chan$^{3}$
\thanks{$^{1}$ F. Ye is with TuSimple Inc., 9191 Towne Centre Dr. Ste 600, San Diego, CA 92122, USA. {\tt\small fei.ye@tusimple.ai.}}
\thanks{$^{2}$ S. Zhang is with Georgia Institute of Technology, Atlanta, GA 30332, USA. {\tt\small shenzhang@gatech.edu.}}%
\thanks{$^{3}$ P. Wang and C. Chan are with California PATH, University of California, Berkeley, Richmond, CA 94804, USA. 
        {\tt\small \{pin\_wang, cychan\}@berkeley.edu.}}%
}
\begin{document}
\maketitle
	
\begin{abstract}
In this survey, we systematically summarize the current literature on studies that apply reinforcement learning (RL) to the motion planning and control of autonomous vehicles. Many existing contributions can be attributed to the pipeline approach, which consists of many hand-crafted modules, each with a functionality selected for the ease of human interpretation. However, this approach does not automatically guarantee maximal performance due to the lack of a system-level optimization. Therefore, this paper also presents a growing trend of work that falls into the end-to-end approach, which typically offers better performance and smaller system scales. However, their performance also suffers from the lack of expert data and generalization issues. Finally, the remaining challenges applying deep RL algorithms on autonomous driving are summarized, and future research directions are also presented to tackle these challenges.
\end{abstract}
\section{Introduction}
Automated and semi-automated vehicles are gaining popularity in assisting our daily transportation. There is a considerable amount of studies in the past decade focusing on autonomous driving applications \cite{gonzalez2015review, classical_survey, schwarting2018planning, yurtsever2019survey, talpaert2019exploring, kiran2020deep}. Specifically, a large number of research activities based on deep learning have been conducted for advanced driving assistance systems (ADAS) and automated driving applications, aiming to automate as much of the driving task as possible. Supervised learning approaches rely heavily on large amounts of labeled data to be able to generalize and it is basically trained on each task in isolation. However, obtaining a big amount of data for each individual task in autonomous driving is costly and time-consuming. Moreover, it requires massive human labor to label such data and still may not cover all the complex situations in the real-world driving.

On the other hand, reinforcement learning (RL) algorithms have been extensively applied to vehicle decision making and control problems \cite{talpaert2019exploring, kiran2020deep}. Specifically, RL is able to learn in a trial-and-error way and does not require explicit human labeling or supervision on each data sample. Instead, it needs a well-defined reward function to receive reward signals in its learning process. Additionally, there is a wide variety of deep RL algorithms and high flexibility in the implementation level, such as the state space, the action space, and the reward function, etc. 

\begin{figure*}[!t]
\centering
\includegraphics[width=\linewidth]{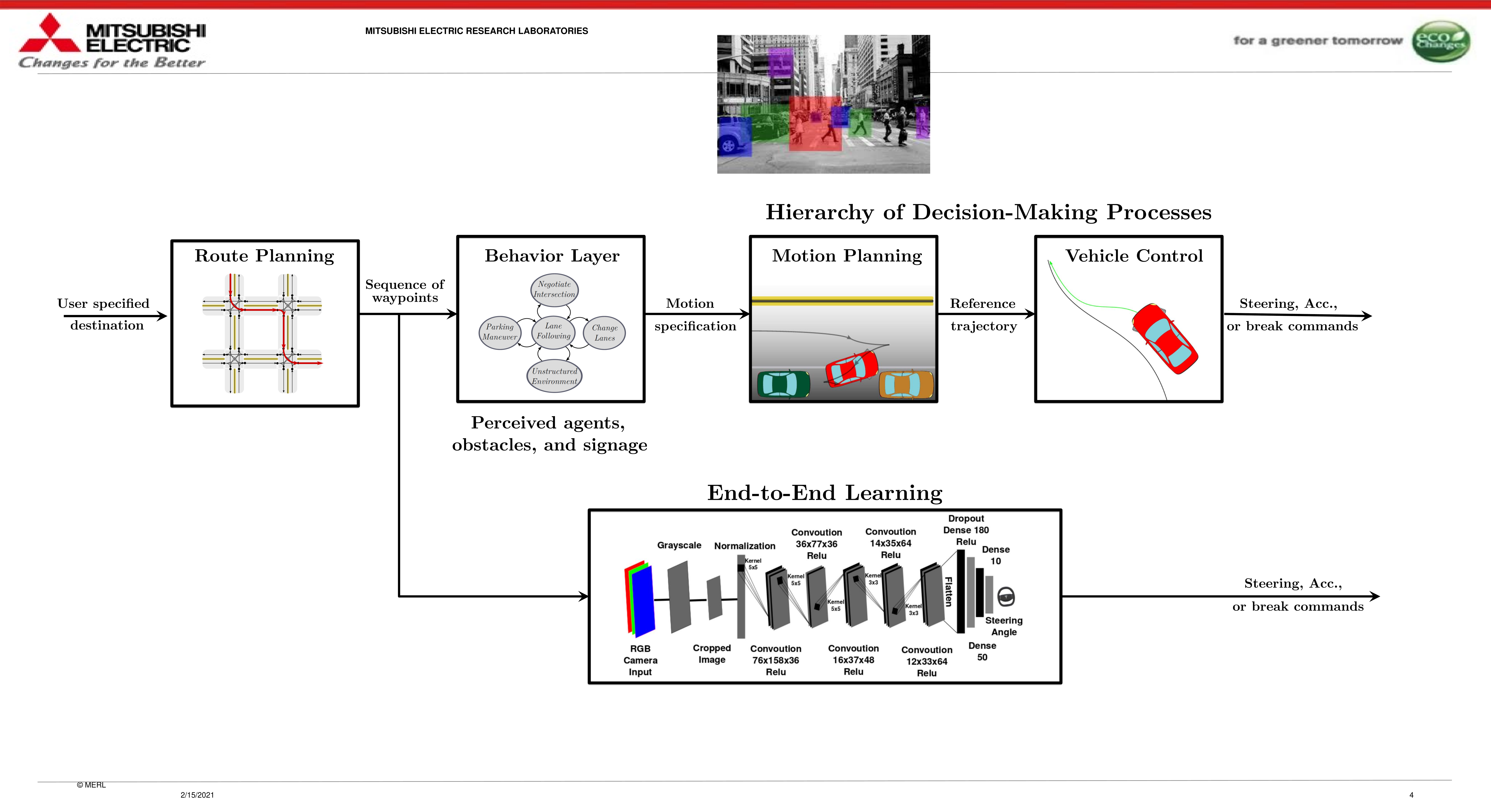}
\caption{An illustration of the pipeline and the end-to-end approach for motion planning and control of autonomous vehicles, figure adapted from \cite{classical_survey}.}
\label{fig_architecture}
\end{figure*}

In general, existing work applying deep RL to the motion planning and control of autonomous vehicles can be divided into the hierarchical (pipeline) approach and the end-to-end approach \cite{aradi2020survey}, as demonstrated in Fig. \ref{fig_architecture}. Specifically, the pipeline approach can be typically categorized into different modules such as perception, decision making, motion planning, low-level control, and so on. In this paper we'll review and examine papers that apply deep RL algorithms to specifically accomplish functionalities of motion planning
and vehicle control.
Each of these modules is engineered manually, and the interfaces between the modules are typically implemented by hands. However, this modular distribution is obviously targeted for the convenience of human interpretation, rather than the highest attainable system performance. For example, if there is a pipeline system with parts that can improve with data, and with parts that they don't, then those parts that do not improve with data will eventually become the bottleneck. 

Recently, more efforts have been devoted to the second approach enabled by deep RL, which is the end-to-end approach that can optimize all those modules of abstractions to map sensory input to control commands with the minimal number of processing steps.
There are mainly two reasons why we might want to apply end-to-end techniques to autonomous driving: 1) better performance; and 2) smaller system scales \cite{Levine_video}. Better performance will result because the internal components can be self-optimized to maximize overall system performance, instead of optimizing human-selected intermediate criteria. Smaller networks are possible because the system learns to solve problems with minimal processing steps.

In this context, this paper seeks to summarize existing work that explains how deep RL algorithms, when combined together with deep neural network representations, can generalize and perform automated vehicle decision making and control. The rest of the paper is organized as follows. In Sections II and III, we introduce how different deep RL algorithms can be leveraged to accomplish behavioral decision making, motion planning and control modules in the pipeline approach. Next, in Section IV, we take a deep dive into end-to-end learning methods for both real-world applications and the sim-to-real approach. Section V summarizes existing challenges and future work directions in applying deep RL to autonomous vehicles. Finally, a conclusion is provided in Section VI.

\section{Deep Reinforcement Learning for Decision Making and Motion Planning}
In this section, we focus on recent advances in behavioral decision-making and motion planning for autonomous vehicles based on deep RL. As shown in Fig. \ref{fig_architecture}, the typical pipelines of an autonomous driving system process a stream of observation from vehicle on-board sensors with high level routing plans to the executable control output such as steering angles, accelerations and braking actions. Typically, a hierarchical structure in the design of planning systems for autonomous driving is usually desired, since driving is naturally hierarchical as higher level decisions are made on discrete state transitions and lower level executions are performed in continuous state space. Behavior layer is a decision making system that determines the transition of discrete state of mid-level driving behaviors such as lane changing, car-following, turning left/right, etc. When the behavior decision is made, the motion planning system is responsible for providing a safe, comfortable, and dynamically feasible continuous trajectory to achieve the selected driving behavior from the decision making system.  

Deep reinforcement learning has shown great success in the area of vehicle behavioral decision makings, especially in the highway scenarios and urban intersections. To reduce the sample complexity, some of the studies choose to adopt the mid-level inputs from the perception system processes and extract the vehicle state and relative states of surrounding vehicles as inputs. Hoel et al \cite{hoel2018automated} trained a Deep Q-Network (DQN) in a simulation environment to issue driving behavioral commands (e.g. change lanes to the right/left, cruise on the current lane and etc.) and compared the different neural network structures' effects on the agent performance. To migrate the concerns of safety performance of a trained RL agent, it is common to add rule-based safety constraints that can verify unsafe actions before they are executed. In \cite{isele2018safe}, a prediction model combined with Deep Q-Network is proposed given the outputs from the perception system to label the unsafe behavioral decisions in unprotected turn scenarios. Some studies  \cite{wang2019lane, mirchevska2018high} alternatively trained a lane change decision making system based on DQN for behavioral level decision-making and uses an underlying rule-based layer to verify the safety of a planned trajectory before it is executed by the vehicle control system. In \cite{shi2019driving}, a hierarchical RL based architecture is presented to combine lane change decisions and motion planning together. Specifically, a deep Q-network (DQN) is trained to decide when to conduct the maneuver based on safety considerations, while a deep Q-learning framework with quadratic approximator is designed to complete the maneuver in longitudinal direction. On the other hand, some of the studies choose to learn from human demonstrations via imitation learning and introduce perturbation to discourage undesirable behavior \cite{Chauffeurnet}. 

More recently, actor-critic policy-based RL methods are introduced in autonomous vehicle decision making and motion planning. Compared to value-based RL methods such as Deep Q-Networks that approximate the value function using neural networks in an off-policy way, the primary advantage of actor-critic policy-based RL method is that they can directly compute actions from the policy gradient rather than optimizing from the value function, while remaining stable during function approximations. On the other hand, the  merit of the critic is to supply the actor with the knowledge of performance in low variance. Actor-Critic algorithm has been successfully applied in both discrete behavior decision makings and continuous motion planning tasks \cite{ye2019automated, ye2020automated}.

\section{Deep Reinforcement Learning for Vehicle Control}
In the pipeline approach, the perception layer such as CNN is typically used to process and recognize objects in digital images. Afterwards, deep RL algorithms can be applied to understand and learn intelligent actions and controls. Many papers are devoted to accomplishing low-level vehicle control with deep RL methods, such as lane keeping \cite{mazumder2018action, ma2018improved}, lateral control \cite{sun2017fast, wang2019continuous, li2019reinforcement}, longitudinal control  \cite{pietquin2011batch, lefevre2015learning, schultz2018deep, zhu2018human, hartmann2019deep, lin2019longitudinal, chen2019model}, or both \cite{shalev2016safe, he2018human, wang2019quadratic}.

For lane-keeping, a DDPG model was implemented in continuous state and action space in \cite{mazumder2018action} to guide its training and apply it to solve the lane keeping (steering control) problem in self-driving or autonomous driving. It is shown that the proposed method can help speed up RL training remarkably for the lane keeping task as compared to the RL algorithm without exploiting the state-action permissibility-based guidance and other baselines that employ constrained action space exploration strategies. To improve efficiency and reduce failures in autonomous vehicles, two different algorithms, namely the robust adversarial RL and neural fictitious self play are proposed in \cite{ma2018improved}, and compares performance on lane keeping and lane changing scenarios. The results exhibit improved driving efficiency while effectively reducing collision rates compared to baseline control policies produced by traditional RL methods.

In terms of the vehicle lateral control, to address the high computational complexity of model predictive control (MPC) for real-time implementation, a fast integrated planning and control framework is proposed in \cite{sun2017fast} that combines a driving policy layer and an execution layer. 
Several example driving scenarios demonstrated that the performance of the policy layer can be improved quickly and continuously online. In \cite{wang2019continuous}, DDPG was implemented to formulate the lane change behavior with continuous action in a model-free dynamic driving environment, and the reward function takes the position deviation status and the maneuvering time into account. Eventually, the RL agent is trained to smoothly and stably change to the target lane with a success rate of 100\% under diverse driving situations in simulation. In \cite{li2019reinforcement}, a vision-based lateral control system was broken into a perception module and a control module, which is based on deep RL to make a control decision based on features coming from the perception module. The trained RL controller in visual TORCS outperforms the linear quadratic regulator (LQR) controller and model MPC controller on different tracks.
\section{End-to-End Deep Reinforcement Learning for Autonomous Driving}
The previous two sections summarize existing work that fall into the pipeline approach, which consists of many hand-crafted modules for the ease of human interpretation, such as perception, motion planning, decision making, low-level control, and so on. However, this pipeline approach does not guarantee maximum system performance if some parts of modules cannot improve with data, which will eventually become the bottleneck. 

On the other hand, the end-to-end approach is expected to have better performance and smaller systems for autonomous driving. With end-to-end training, it is not possible to make a clean break among different modules of a system, such as which parts of the network mainly act as feature extractors and which ones serve as controllers. However, it is able to optimize all those modules of abstractions simultaneously with a reduced number of processing steps, which indicates that the abstractions will be optimally and automatically adapted for the task that needs to be solved, instead of optimizing human-selected intermediate criteria. 
\subsection{Joint Optimization in Real-World Applications}

A very early attempt that successfully steered a car on public roads using the end-to-end approach can be found in Pomerleau's pioneering work in 1989 \cite{alvinn}, in which controlling an autonomous land vehicle was trained using a neural network system. In the early 2000's, LeCun et al. built a small off-road robot that uses an end-to-end learning system to avoid obstacles solely from visual input \cite{DAVE}. The robot named DAVE was trained on images sampled from human driving videos, paired with the corresponding steering command $1/r$. While DAVE demonstrated the potential of end-to-end learning, its performance was not reliable enough to provide a complete alternative to more modular methods of off-road driving, as its mean distance between crashes was around 20 meters in complex environments. 

With the evolution of hardware computation capabilities and the advancement of modern deep learning algorithms, in 2016, researchers at NVIDIA proposed an end-to-end learning based on CNN that learns the entire processing pipeline needed to directly steer an automobile \cite{NVIDIA}. Training data was collected from less than a hundred hours of expert driving on a wide variety of roads, paired with time-synchronized steering commands generated by the human driver. The sampled images are fed into a CNN consisting of five convolutional layers with three fully connected layers, which then computes the proposed steering command. The weights of the CNN are optimized by minimizing the mean squared error between the proposed command and the benchmark command for that image. This milestone also empirically validated that CNNs can ``learn the entire task of lane and road following without manual decomposition into road or lane marking detection, semantic abstraction, path planning, and control'' \cite{NVIDIA}. 

Direct deep supervised learning usually requires a large amount of data to learn a generic driving policy and the ground truth labeling for training. For example, \cite{xu2017end} proposes a novel FCN-LSTM architecture to leverage both previous vehicle states and current visual observations using a long short-term memory temporal encoder with a fully convolutional visual encoder. Meanwhile, this work also released the Berkeley DeepDrive Video dataset (BDDV) for learning driving models and used the human driving behavior as the ground truth label for training. As such, human labeling can be costly and time-consuming. Furthermore, the policy is trained to mimic human behavior in a certain scenarios and it is hard to cover all the real-world scenarios.

By contrast, a full end-to-end reinforcement learning approaches learn policy in a trail-and-error way, and would not require such supervision. As presented in Sections II and III, deep RL algorithms have been widely employed as independent motion planning or control modules for autonomous vehicles. If extended, they can be also trained together with convolutional layers to constitute an end-to-end approach via joint optimization. For instance, using a single monocular image as input, the actor-critic algorithm was adopted in \cite{kendall2019learning} to learn a policy for lane following in a handful of training episodes, which is trained to maximize the reward of distance that an agent can travel before intervention by a safety driver. Similarly, an asynchronous actor critic (A3C) framework established in \cite{game2} was used to learn vehicle control, and a thorough evaluation was conducted on unseen tracks and using legal speed limits. This work also demonstrated good domain adaptation capability when testing the proposed control commands on real videos.

Besides the aforementioned end-to-end on-road driving tasks that need to generalize to a larger domain and contend with moving objects such as cars and pedestrians, an end-to-end learning system for agile, off-road autonomous driving using only low-cost on-board sensors is presented in \cite{pan2017learning}. Compared with paved roads, the surface of our dirt tracks are constantly evolving and highly stochastic. As a result, to successfully perform high-speed driving in our task, high-frequency decision and execution of both steering and throttle commands are required. By imitating an optimal controller, a deep neural network control policy was successfully trained to map raw, high-dimensional observations to continuous steering and throttle commands. 

\subsection{Simulation}
Despite the remarkable advancement of end-to-end learning in real-world applications, there are still many challenges that hinders their on-road deployment in full autonomy. Specifically, many of these challenges arise from the fact that rarely are these vehicles trained or tested at all possible scenarios (including corner cases). However, manually creating these cases and collecting data in the real world can be a process that is expensive and often impractical. 

Therefore, a very promising approach is to leverage simulation to gain experience in these situations. In simulation, the cost of putting an obstacle on the road or simulating a car crash is negligible, but it still offers valuable experiences, whilst in a real-world scenario, such unexpected events can lead to severe financial losses and even threaten the lives of pedestrians or people in neighboring vehicles. 
\begin{figure}[!t]
\centering
\includegraphics[width=3.3in]{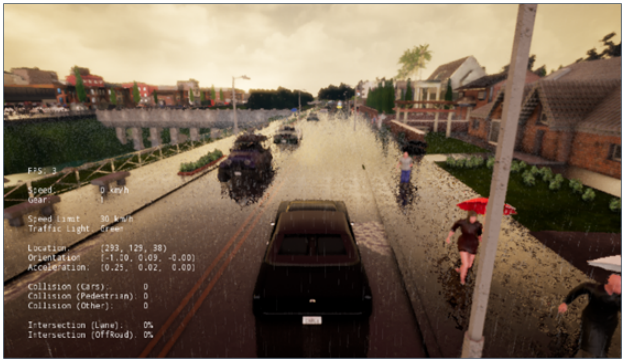}
\caption{Carla environments: Hard Rainy in Town 1 \cite{jaafra2019robust}.}
\label{fig_Carla}
\end{figure}
While simulation has long been an essential part of testing autonomous driving systems, such as Carcraft of Waymo \cite{Waymo} and Drive Constellation of NVIDIA \cite{Constellation}, only recently has simulation been applied to building and training end-to-end self-driving vehicles on driving simulation platforms, including TORCS \cite{koutnik2013evolving, lillicrap2015continuous, zhang2016query, pan2017virtual, wang2018deep, chhorrobust}, CARLA \cite{IL1, liang2018cirl, Rhinehart, Rhinehart2, jaafra2019robust}, Unity \cite{klose2019simulated}, WRC6 \cite{game1, game2}, and Vdrift \cite{carma}. 

As a highly portable multi platform car racing simulation, TORCS \cite{TORCS} can be used as an ordinary car racing game, but also as a research platform. To our knowledge, the first attempt to control vehicles on TORCS using end-to-end technique is presented in \cite{koutnik2013evolving} for an RL competition \cite{rl_competitionn} using vision from the driver’s perspective, which is later processed with CNN and RNN. Since then, many different deep RL algorithms have been applied to drive a vehicle end-to-end on this platform, such as deep deterministic policy gradient (DDPG) \cite{lillicrap2015continuous, wang2018deep, chhorrobust}, SafeDAgger \cite{zhang2016query}, and A3C \cite{pan2017virtual}, etc. Specifically, \cite{lillicrap2015continuous} is the original paper that proposed the DDPG algorithm to handle complex state and action spaces in continuous domain, which is extremely useful to generate continuous action spaces that can adapt to the complex real-world driving scenarios.

Both CARLA \cite{CARLA} and Unity \cite{Unity} can be used to efficiently create more realistic simulation environments that are rich in sensory and physical complexity. Their major difference lies in CARLA being an explicit driving simulation while the basic Unity being a more generic engine, and does not describe a very specific realization. CARLA has been developed to support development, training, and validation of autonomous driving systems, which also provides an interface allowing an RL agent to control a vehicle and interact with a dynamic environment. An illustration of the CARLA environment with Hard Rainy in Town 1 is presented in Fig. \ref{fig_Carla}. Since a vehicle trained end-to-end to imitate an expert cannot be controlled at test time (i.e., cannot take a specific turn at an upcoming intersection), a condition imitation learning approach was proposed in \cite{IL1} to enable the learned driving policy to behave as a chauffeur that handles sensorimotor coordination but also continues to respond to navigational commands. To alleviate the the inefficiency of exploring large continuous action spaces that often prohibits the use of classical RL in challenging real driving tasks, \cite{liang2018cirl} proposed a controllable imitative RL based on DDPG to explore over a reasonably constrained action space guided by encoded experiences that imitate human demonstrations. Extensive experiments on CARLA demonstrate its superior performance in terms of the percentage of successfully completed episodes and good generalization capability in unseen environments. Similarly, to achieve a better robustness of the agent learning strategies when acting in complex and unstable environments, an advantage actor-critic algorithm was implemented in \cite{jaafra2019robust} with multi-step returns.

\begin{figure}[!t]
\centering
\includegraphics[width=\columnwidth]{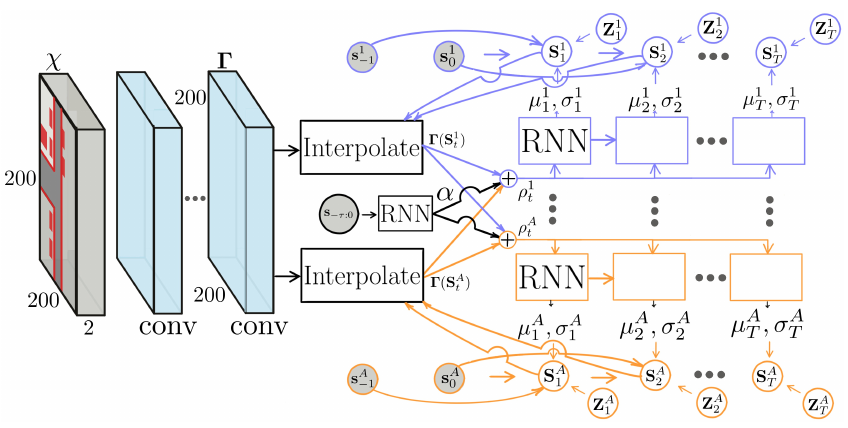}
\caption{An autonomous driving model using combined deep imitation learning and model-based reinforcement learning \cite{Rhinehart2}.}
\label{fig_imitative_model}
\end{figure}
Another innovative deep RL algorithm that is worth mentioning is introduced by Rhinehart et al. in \cite{Rhinehart} and \cite{Rhinehart2}. Specifically, the proposed imitative model, as depicted in Fig. \ref{fig_imitative_model}, combines imitation learning with model-based RL to develop probabilistic models to overcome the difficulty of specifying appropriate reward functions that are crucial to evoke desirable behaviors. Relying on LiDAR data inputs, the effectiveness of the proposed imitative model to predict expert-like vehicle trajectories is validated using CARLA, but without including other vehicles and pedestrians.
\begin{table*}
\centering
    \caption{A Summary of Different Deep RL Formulations on the Motion Planning of Autonomous Vehicles.}
    \resizebox{\linewidth}{!}{
\begin{tabular}{lcccc}
\hline
\rowcolor[HTML]{C0C0C0} 
                & \textbf{Algorithm}       & \textbf{State} $s_t \in S$ & \textbf{Action} $a_t \in A$  & \textbf{Reward} $R$ / \textbf{Loss} $\ell$\\
\hline
\rowcolor[HTML]{EFEFEF} 
\textbf{Motion Planning (Pipeline)} &          &          &          &        \\
\hline
%
%
Hoel et al. (2018) \cite{hoel2018automated}  &   DQN  &   Vehicle \& relative states &  $A=\{\mathrm{Lane~change, Acc., Brake}\}$   &   $R\{\mathrm{+:Efficiency, Comfort, Safety}\}$ \\ 
Isele et al. (2018) \cite{isele2018safe}  &   Classical RL   &   Local states &  $A=\{\mathrm{Safe~driving~decision}\}$   &   $R\{\mathrm{+:Safety}\}$ \\ 
Bansal et al. (2018) \cite{Chauffeurnet}  &   Imitation learning   &   Traffic light/dynamic object states  &  $A=\{\mathrm{Driving~trajectory}\}$    &  9 training losses  \\
Wang et al. (2019) \cite{wang2019lane}  &   DQN   &   Vehicle states  &  $A=\{\mathrm{Lane~change~decision}\}$    &   $R\{\mathrm{+:Speed; -: Collision, Invalid~Decision}\}$  \\
Mirchevska et al. (2018) \cite{mirchevska2018high}  &   DQN   &   Vehicle \& relative states  &  $A=\{\mathrm{Lane~change~decision}\}$    &   $R\{\mathrm{+:Speed}\}$  \\
Ferdowsi et al. (2018) \cite{ferdowsi2018robust}  &   Adversarial RL   &   Vehicle states  &  $A=\{\mathrm{Optimal~safe~action}\}$    &   $R\{\mathrm{+:Safety (distance)}\}$  \\
Ye et al. (2019) \cite{ye2019automated}  &   DDPG   &   Vehicle \& relative states  &  $A=\{\mathrm{Lane~change~decision}\}$    &   $R\{\mathrm{+:Efficiency, Comfort, Safety}\}$  \\
Shi et al. (2019) \cite{shi2019driving}  &   DQN   &   Vehicle relative states &  $A=\{\mathrm{Lane~change~decision}\}$   &   Hand-Crafted $R(s_t,a_t)$ \\  
Nishi et al. (2019) \cite{nishi2019merging}  &   MPDM \& pAC   &   Vehicle relative states &  $A=\{\mathrm{Lane~merging~trajectory}\}$   &   Hand-Crafted $R(s_t,a_t)$ \\  
You et al. (2019) \cite{you2019advanced}  &   Deep inverse RL   &   Grid-form states &  $A=\{\mathrm{Optimal~driving~strategy}\}$   &   Hand-Crafted $R(s_t,a_t)$ \\  
Bouton et al. (2019) \cite{bouton2019reinforcement}  &   Q-learning   &   Vehicle \& agent states &  $A=\{\mathrm{Strategic~maneuvers}\}$   &   $R\{\mathrm{+:Goal, Efficiency; -: Collision}\}$ \\
Ye et al. (2020) \cite{ye2020automated}  &   PPO   &   Vehicle \& relative states  &  $A=\{\mathrm{Lane~change~decision, Car~ following}\}$    &   $R\{\mathrm{+:Efficiency, Comfort, Safety}\}$  \\
\hline
\rowcolor[HTML]{EFEFEF} 
\textbf{End-to-End}      &                 &             &         &        \\
\hline
LeCun et al. (2005) \cite{DAVE}     &    Imitation learning   &   Camera image   &  $A=\{a|a\in [-\max Left, +\max Right]\}$    &  $\ell^2~\mathrm{loss}$    \\
Lillicrap et al. (2015) \cite{lillicrap2015continuous}        &   DDPG      &  Simulator image     & $A=\{\mathrm{Steering, Acc., Brake}\}$     &   $R\{\mathrm{+1:Direction; -1:Collision}\}$     \\
Bojarski et al. (2016) \cite{NVIDIA}     &    Imitation learning   &   Camera image   &  $A=\{1/r|r\in \mathrm{\{turning~radius\}}\}$    &  $\ell^2~\mathrm{loss}$    \\
Vitelli et al. (2016) \cite{carma}     &   DQN   &   Images and estimated states   &  $A=\{\mathrm{Steering, Acc., Brake}\}$    &  Hand-Crafted $R(s_t,a_t)$ \\
Xu et al. (2016) \cite{xu2017end}     &   FCN-LSTM   &   Visual and sensor states   &  \begin{tabular}[c]{@{}c@{}}Disc. $A=\{\text{Straight, Stop, Left, Right}\}$\\ Continuous $A=\left\{\vec{v}|\vec{v} \in \mathbb{R}^{2}\right\}$\end{tabular}    &  Cross entropy loss    \\
Zhang et al. (2016) \cite{zhang2016query}     &   SafeDAgger   &   Camera image   &  $A=\{\mathrm{Steering, Acc., Brake}\}$    &  $R\{\mathrm{+1:No~crash}\}$    \\
Perot et al. (2017) \cite{game1}     &   A3C   &   Game states   &  $A=\{\mathrm{Steering, Acc., Brake}\}$    &  $R=v(\cos \theta-d)$  \\
Codevilla et al. (2017) \cite{IL1}     &   Imitation learning   &  Images \& driver internal states   &  $A=\{\mathrm{Steering, Acc.}\}$    &  $\ell^2~\mathrm{loss}$  \\
Pan, Cheng et al. (2017) \cite{pan2017learning}     &   Imitation learning   &   Camera image \& vehicle speed  &  $A=\{\mathrm{Steering, Throttle}\}$    &  Steering/Throttle loss  \\
Pan, You et al. (2017) \cite{pan2017virtual}  &   A3C   &   Virtual world images  &  $A=\{\mathrm{Steering, Acc., Brake}\}$    &  $R=\{+v(\cos \alpha-d),-\mathrm{Collision}\}$  \\
Wang et al. (2018) \cite{wang2018deep}  &  DDPG   &   LiDAR sensor \& camera image  &  $A=\{\mathrm{Steering, Acc., Brake}\}$   &  \begin{tabular}[c]{@{}c@{}}$R_{t}=V_{x} \cos (\theta)-\alpha V_{x} \sin (\theta)$\\ $~~~~~~-\gamma|P o s|-\beta V_{x} |P o s |$ \end{tabular}   \\
Kendall et al. (2018) \cite{kendall2019learning}  &   DDPG   &   Camera image   &  $A=\{\mathrm{Steering, Speed}\}$    &  Distance travelled without infraction  \\
Jaritz et al. (2018) \cite{game2}  &   A3C   &   Input image   &  $A=\{\mathrm{Steering, Acc., Brake, Hand~brake}\}$    &  $R\{\mathrm{+1:On~track; In~lane}\}$  \\
Klose et al. (2018) \cite{klose2019simulated}  &   DQN   &   Raw sensor input   &  $A=\{\mathrm{Acc., Brake}\}$    &  $R=\sum_{i=1}^{3}\exp \left(-0.5 \cdot\left[{x_{t}^{i}}/{\theta_{i}}\right)^{2}\right]$ \\
Liang et al. (2018) \cite{liang2018cirl}  &   Imitative RL   &   Human driving videos   &  $A=\{\mathrm{Steering, Acc., Brake}\}$    &  $R\{\mathrm{+:Speed; -1:Collision, Steering}\}$ \\
Rhinehart et al. (2018) \cite{Rhinehart}  &   Imitative model   &   LiDAR image  &  $A=\{\mathrm{Driving~way~points}\}$    &   Probabilistic inference objectives  \\
Min et al. (2019) \cite{min2019deep}  &  DDPG   &   Raw sensor input  &  $A=\{\mathrm{Lane~change, Acc., Brake}\}$   &  $R\{\mathrm{+1:Speed; -1:Collision}\}$   \\
Chhor et al. (2019) \cite{chhorrobust}  &   DDPG   &   Raw sensor input   &  $A=\{\mathrm{Steering, Acc., Brake, Hand~brake}\}$    & $R=V \cos \theta-V \sin \theta-V|trackPos|$ \\
Jaafra et al. (2019) \cite{jaafra2019robust}  &   A2C   &   Camera image   &  $A=\{\mathrm{Steering, Throttle, Break}\}$    & $R\{\mathrm{+1:Closing~goals;-1: Not~in~lane}\}$  \\
Rhinehart et al. (2019) \cite{Rhinehart2}  &   Imitative model   &   LiDAR image  &  $A=\{\mathrm{Driving~way~points}\}$    &   Probabilistic inference objectives  \\
\hline
\end{tabular}}
    \label{tab:DL_sum1}
\end{table*}
\subsection{The Sim-to-Real Approach}
\begin{figure}
\centering
\includegraphics[width=\columnwidth]{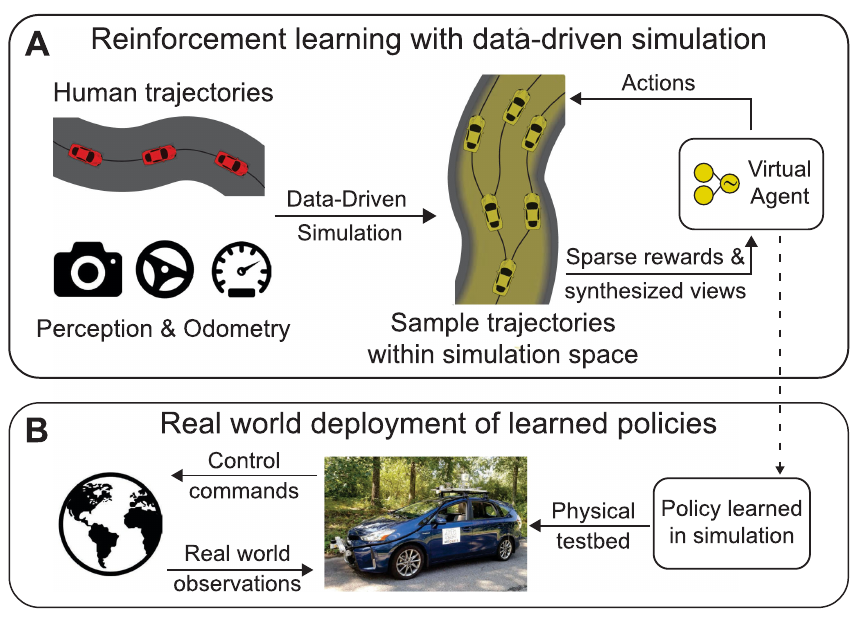}
\caption{Training and deployment of policies from Sim-to-Real transfer \cite{Sim-to-Real_MIT}.}
\label{fig_Sim_to_Real}
\end{figure}

%
By leveraging the simulated driving scenarios and experiences obtained via simulation, the sim-to-real approach is a good alternative to train an end-to-end driving policy without using real data, as shown in Fig. \ref{fig_Sim_to_Real}. This field has received a lot of attention during the recent years, and it has been shown in \cite{sadeghi2016cad2rl} that just by randomizing the simulator very carefully, a policy can be trained to fly drones indoors using the simulated data without having to do very careful system identification.

Specifically, the simulator can facilitate the training of deep neural networks by generating abundant labeled data in many corner cases to extract task-relevant features and acquire good state representations, and the learned knowledge is expected to promote faster learning and better performance in real world scenarios. Therefore, the sim-to-real transfer is an important area of research that can adapt the learned knowledge from vehicle-traffic simulations to the real-world environment for decision making, planning and control.

However, driving autonomously in the urban environment consists of multiple tasks that involve complex and uncertain driving behaviors and interactions with the surrounding traffic. Besides some typical tasks that are shared with highway driving, such as lane keeping, lane changing, overtaking, and car following, driving in the urban environment also includes taking left-turns, complying with road signs and traffic lights, keeping an eye on lower-speed pedestrians and bicyclists, etc. While each of these specific tasks can be separately modeled in the simulator to train the autonomous driving policy with an outstanding performance, the knowledge transfer from the simulator to the real-world scenario would be more challenging due to the large number and complexity of tasks that further intensifies the difference between the source domain (simulator) and the target domain (real-world). Moreover, there are other technical challenges such as real-world visual signal
noises, and the training environment in a car driving simulator is often significantly different from real-world
driving in terms of their visual appearance \cite{pan2017virtual}.

Due to the aforementioned challenges, while many research work applied some variant of deep RL algorithms on the simulation platform, only a handful of them attempted to transfer the knowledge learned from the simulator to real-world applications, i.e., \cite{game2}. In fact, \cite{game2} claimed itself to be ``the first time of a deep RL driving'', which is trained using the simulator, ``is shown working on real images'', and foresees simulation based RL can be used as initialization strategy for networks used in real applications. However, it is also reported in \cite{Sim-to-Real_MIT} that ``end-to-end models trained solely in CARLA were unable to transfer to the real world'', so some domain randomization and domain adaptation methods are needed for bridging the sim-to-real gap in simulators.
\subsection{Summary}
For the purpose of providing a comprehensive and intuitive comparison among different deep RL formulations on the motion planning and control of autonomous vehicles, TABLE \ref{tab:DL_sum1} summarizes the specific algorithm, the state space, the action space, and the specific reward/loss design of selective papers included in this review.
\section{Challenges and Future Work Directions}
\subsection{Challenges}
\subsubsection{The Pipeline Approach}

The choice of hand-crafted abstractions (features) for each module can limit the performance of the entire system. Once designed, these abstractions will have limited capacity to improve, and those parts that do not improve with data will eventually become the bottleneck. For instance, if the perception system becomes better, but the planner doesn't get any better to utilize those benefits, eventually the planner will be the bottleneck. Ultimately we don't actually know how to accurately construct the correct abstractions for the real world, and eventually these will get us into trouble. 

One potential approach to tackle this challenge is to train all those layers of abstractions end-to-end, which means that the abstractions were optimally adapted for the task that needs to be solved.  
\subsubsection{The Sim-to-Real Approach}
Generally speaking, training with data from only the human driver is not adequate, and collecting a sufficient amount of data from every possible driving condition can be extremely expensive and dangerous. While the sim-to-real approach can help people get away with no real-world data at all, this approach might not be sufficient to solve the problem, even though it is still an excellent way to get the network and parameters initialized. This is because instead of designing each of these pipeline modules by hand, we'll still have to design our simulator by hand. In the end, the simulator will become the bottleneck due to challenges for developing a sufficiently realistic simulator with diverse environments \cite{Sim-to-Real_review}, including the modeling of any interacting traffic participant with realistic dynamics. 

\subsection{Future Work Directions}
Despite the exciting progress of applied deep RL algorithms in autonomous driving tasks, these approaches are mostly trained for one task at a time, and each new task requires training a new agent, which is data-inefficient and fails to exploit the learned properties of similar tasks. In contrast, humans have the ability to not only learn complex tasks, but they can also adapt rapidly to new or evolving situations. 

Therefore, some methods have been proposed to increase the generalization of deep RL algorithms in a variety of driving tasks. 
On the one hand, randomizing the simulator environment has been applied to generalize the trained policies \cite{2017domain}. On the other hand, transferring the knowledge accumulated from past experience through continual learning and meta learning has recently been explored to facilitate the generalization in both reinforcement learning \cite{ye2020meta} and imitation learning \cite{meta_framework_2017}. The ability to continuously learn and adapt quickly is essential to achieving real-world automated driving, which motivates further studies to introduce transfer learning and meta learning concepts into deep RL in the realm of autonomous driving. 
\section{Conclusions}
In this paper, a systematic review is conducted on the existing literature employing deep RL algorithms on motion planning and control of autonomous vehicles, which is a field that has spurred the interest of industry and academia over the past five years. Both pipeline and end-to-end approaches have been extensively discussed. It has been demonstrated that, despite the fact that deep RL algorithms require an extended period and a large dataset to train, they can effectively interact with the environment in a trial-and-error way and does not require explicit human labeling or supervision on each data sample, making them promising candidates to accomplish autonomous driving in real-world applications. 

%



%
%
\balance
\bibliographystyle{IEEEtran}
\bibliography{IEEEabrv.bib,ref.bib} 

\end{document}